# What Does Matching a Human Intervention Effect Establish? Benchmark Checks for LLM Simulations

Tengfei Shao, PhD[1], Asma Rezaei, PhD[2], and Ali Reza Kamali, PhD[3]

[1] Global Education Center, Waseda University, Tokyo, Japan

[2] Faculty of Business and Law, School of Management, Anglia Ruskin University, Cambridge CB1 1PT, United Kingdom

[3] Cambridge Silicon Age Centre for Innovation, Barnwell Business Park, Cambridge CB5 8UZ, United Kingdom

Corresponding author: Tengfei Shao, Global Education Center, Waseda University, 1-6-1 Nishiwaseda, Shinjuku-ku, Tokyo 169-8050, Japan. Email: tengfei.shao@toki.waseda.jp

ORCID: Tengfei Shao, https://orcid.org/0000-0003-3089-5696; Ali Reza Kamali, https://orcid.org/0000-0002-2849-8547

**Abstract**

LLM simulations are validated by equivalence with human effects, yet a zero reference passes whenever the human 90 percent interval lies inside the margin, as the human benchmark shows before simulation. When the zero reference fails, planned precision gives independent simulations the margin up to which passing is a 5 percent test of response. In an exploratory 70-experiment reanalysis, 72 percent of GPT-4's ±0.20 SD passes were zero-reference passes; GPT-4 passed 16 contrasts the zero reference failed (1.58 expected without response). In a preregistered audit, Claude Sonnet 5.5 passed three intention targets, all zero-reference passes, and the registered belief-gradient control; norm beliefs changed 0.065 SD against humans' 0.351 (zero-reference check added after unblinding; norm beliefs emphasized after inspection). Registered tests detected no overestimation; average-effect intervals excluded overestimation above 0.08 SD. A group audit's precision floor is 1.33 facts against a 2.9-fact effect. Report the zero-reference threshold beside every equivalence decision.

## 1. Introduction

A large language model (LLM) simulation used to screen interventions, plan human samples, or measure simulated group behavior needs validation against human data (Kozlowski and Evans 2025; Davidson and Karell 2025; Hullman et al. 2026). Effect-level validation can test whether the simulated effect matches the human effect within a prespecified equivalence margin. A reference fixed at zero, which ignores treatment (the zero reference), passes the same test whenever the human estimate's 90 percent interval lies inside the margin. On those benchmarks, a simulation's equivalence pass repeats the verdict that the human benchmark fixed for the zero reference before simulation.

Published evaluations offer agreement with human results as evidence that a simulation works. They replicate experiments with simulated participants (Aher, Arriaga, and Kalai 2023; J. S. Park et al. 2026) and predict experimental results (Ashokkumar et al. 2026). They also compare the significance of simulated and original effects, with significant simulated effects appearing for many nonsignificant originals (Cui, Li, and Zhou 2025). When such agreement guides intervention screening or human-sample planning, its interpretation determines which simulations are trusted.

In the archive that Ashokkumar et al. (2026) released for 70 survey experiments, GPT-4's predicted and observed effects correlated at $r = .81$ across the 156 hypothesized contrasts. Unscaled predictions passed the equivalence criterion on 58 of them at ±0.20 SD, against 61.4 expected from a nonresponsive noise reference, a simulation with true effect zero and GPT-4's prediction noise. Of the 58 passes, 42 (72 percent) were also zero-reference passes, against about half for an oracle reference that reproduces every human estimate up to that noise. A simulation separates from the zero reference when it passes and the zero reference fails. GPT-4 separated on 16 contrasts, against 1.58 expected from the nonresponsive noise reference. The pass count fell 3.4 short of the nonresponsive noise reference's expectation, and the 95 percent interval rules out an excess of more than 9 contrasts, whereas the separation count exceeded its expectation.

This article reads the precision of the human benchmark before simulation and turns the equivalence criterion (TOST, Section 2.1) into a benchmark check. Prediction-powered inference (PPI; Angelopoulos et al. 2023) uses unit-level predictive information, whereas the check uses only the benchmark's effect estimate, standard error, and margin (Section 7.2). Hullman et al. (2026) distinguish validation targets by claim. Balance tests (Hartman and Hidalgo 2018) and the assay sensitivity of active-control trials (Temple and Ellenberg 2000) tie an equivalence decision's meaning to the precision of the data. Belief-updating and deliberative-polling diagnostics evaluate response to information (Pohl et al. 2026; Wali and Tayyab 2026). The article builds on methodological work on generative AI (Davidson and Karell 2025) and, in Study 2, on measurement (Chae and Davidson 2026; Pangakis and Wolken 2025).

The article makes three contributions. First, a benchmark-level check uses a published estimate, standard error, and margin to give the zero-reference threshold and precision floor before simulation. The zero-reference threshold is the minimum supported margin (Hartman and Hidalgo 2018) applied to a fixed zero prediction. Read before simulation, it bounds what an equivalence pass can show. The precision floor is the smallest margin an independent comparison can attain. The same zero reference is the baseline for margin-free scores: significance agreement and mean squared error (MSE) skill (Section 3). The check shows that, in an independent comparison with a true human effect inside the margin, the zero reference passes at least as often as any nonresponsive simulation, averaged across human and simulated samples (Section 2.3). Second, before simulation, an independent comparison of planned precision gives the nonresponse separation probability $P_0$, the probability that a nonresponsive simulation separates from the zero reference. It also gives the margin $\delta_{.05}$ up to which separation is a 5 percent test of response (Section 2.3). Third, a five-step protocol with code pairs each claim with the response or measurement evidence it requires and is applied to the archive and the two studies (Section 6).

Two audits show what evidence the check calls for. Study 1 is a preregistered audit reconstructing two published randomized social-norm experiments, Andre et al. (2026) and Geiger et al. (2025), hereafter Andre and Geiger. The human benchmarks alone showed zero-reference passes on all four registered intention-equivalence targets. Claude Sonnet 5.5 passed three of them and the registered belief-gradient control, which the zero reference fails. Sonnet's mean change on the six-item norm-belief index was 0.065 SD against the human 0.351. Effect agreement, response direction, and response size address three different claims. Study 2 audits a hidden-profile discussion experiment with an LLM facilitator. The benchmark's standard error fixes a 1.33-fact precision floor against a 2.9-fact human effect. A coding audit shows why a measurement check must assess error in the arm contrast rather than in pooled agreement.

## 2. What Equivalence Can Distinguish

The human benchmark alone decides whether the zero reference passes an equivalence criterion.

### 2.1. Target Quantity and Margin

A validation claim names its target quantity, the estimand (Lundberg, Johnson, and Stewart 2021): an average effect, a treatment-effect gradient (the change in effect per unit of a pre-treatment characteristic), a related outcome, or a planning input. Naming it makes the claim specific enough to evaluate and fits functional uses of simulation that preserve selected properties of human responses (Lin 2025).

Let $E$ be the effect error, the simulated estimate minus the human estimate on the outcome's original scale. Two one-sided tests (TOST) at the 5 percent level declare equivalence within a prespecified margin $\pm\delta$ when the 90 percent interval for $E$ lies inside $\pm\delta$ (Schuirmann 1987; Lakens 2017; Lakens, Scheel, and Isager 2018). Political methodology uses equivalence tests for negligible-effect arguments (Rainey 2014). The margin comes from the intended use and is fixed in advance in the units of the compared quantity (the registration's tolerance). Here, response means a nonzero true value of the prespecified effect statistic. A zero average effect can coexist with individual or subgroup responses that cancel. The zero reference and a nonresponsive simulation have a true effect of zero.

### 2.2. The Zero Reference and Separation at δ

The zero reference fixes the simulated coefficient at the known value zero, with no estimation variance. Its error relative to the human coefficient $b_H$ is therefore $-b_H$, with standard error $SE_H$. Under the normal approximation it passes a symmetric $\pm\delta$

TOST criterion exactly when the human study's own equivalence test against zero passes (online supplement Appendix B.1 and Figure B1).

**Decision rule (separation at δ).** A simulation with error $E$ and standard error $SE_E$ passes the same criterion when $|E| + 1.645\ SE_E < \delta$. We call $|E| + 1.645\ SE_E$ the simulation threshold. Two conditions follow, the first from the human estimate alone and so fixed before simulation:

(1) Zero-reference pass: $|b_H| + 1.645\ SE_H < \delta$. We call $|b_H| + 1.645\ SE_H$ the zero-reference threshold.

(2) Separation at δ: $|E| + 1.645\ SE_E < \delta \leq |b_H| + 1.645\ SE_H$, so the simulation passes and the zero reference fails.

Separation therefore requires a zero-reference failure. When both pass, the simulation shares its decision with the zero reference. When $SE_E \geq SE_H$, as in every independent comparison, some margin separates only if $|E| < |b_H|$. The simulation's error must be smaller than the human effect itself. A reversed sign or a simulated effect at least twice the true human effect therefore leaves no margin of separation in population terms (online supplement Appendix B.1.3). With equal standard errors the condition is also sufficient, and condition (2) decides whether the prespecified δ is such a margin.

For independent estimates, $SE_E = \sqrt{(SE_\Delta^2 + SE_H^2)}$, where $SE_\Delta$ is the standard error of the simulated effect Δ. In a joint comparison, such as a stacked estimator on shared respondents, $SE_E$ includes the covariance of the two estimates. More simulated units shrink $SE_\Delta$ while $SE_H$ remains, so every independent comparison needs a margin above the precision floor $1.645\ SE_H$. For a two-arm human benchmark in SD units with equal allocation, the floor falls below δ once the benchmark has more than about $2(1.645/\delta)^2$ participants per arm, about 540 at ±0.10 SD and 136 at ±0.20 SD. This precision condition is necessary for equivalence.

Separation therefore occurs in the informative range, from the precision floor to the zero-reference threshold, whose width is $|b_H|$. The zero reference fails when the human effect reaches the margin, whatever the precision. It also fails when the benchmark is imprecise: $SE_H \geq (\delta - |b_H|)/1.645$ for $|b_H| < \delta$. This imprecision raises the floor that every independent simulation must clear.

### 2.3. Dominance and the Nonresponse Separation Probability

**Proposition (dominance).** Under the normal approximation with fixed standard errors, in an independent comparison with a true human effect β inside ±δ, the zero reference passes at least as often, on average across human and simulated samples, as any nonresponsive simulation. The reason is that a compared difference whose expected value lies inside ±δ has a pass probability that is nonincreasing in its standard error. A nonresponsive simulation has expected error −β and, in an independent comparison, a standard error for the difference of at least $SE_H$. The zero reference is the nonresponsive simulation with no simulation variance. For a true human effect outside the margin, the TOST level bounds every nonresponsive simulation at .05 (online supplement Appendix B.1 gives the proof).

Conditional on the observed human benchmark and the planned simulation precision, and under the same normal, fixed-standard-error assumptions, the nonresponse separation probability $P_0$ is the probability that a nonresponsive simulation satisfies the separation rule. Because the benchmark is fixed before simulation and independent of the simulated estimate, $P_0$ is the exact conditional size of separation read as a test of response (online supplement Appendices B.1.4 and E.2). When the zero reference passes, $P_0 = 0$. The margin $\delta_{.05} = |b_H| + 1.645(SE_E - SE_\Delta)$ is the largest margin at which separation implies that the simulation's own one-sided 5 percent test rejects in the direction of the human effect. Up to $\delta_{.05}$, separation is a 5 percent test of response.

Averaged over repeated human and simulated estimates (online supplement Table B7, analytic, added after unblinding), in an independent comparison with $SE_H = 0.15\delta$ and $SE_\Delta = 0.10\delta$, a simulation whose true effect equals the human effect passes equivalence with probability 1.000. Read as a test of response, it separates with probability .046 when that effect is half the margin and .950 when it equals the margin.

An example from online supplement Table E2 takes $\delta = 0.20$, $SE_H = 0.05$, $b_H = 0.118$, and planned $SE_E = 0.051$. Here, $P_0 = 0.43$ warns before simulation that a nonresponsive simulation would separate at 0.20 in 43 percent of repeated simulations. Up to $\delta_{.05} = 0.185$, below the margin, separation is a 5 percent test of response.

### 2.4. What Equivalence and Response Controls Each Bound

A response control is a prespecified directional test on the model's own outputs, on the outcome and statistic that the response claim names. The registered belief-gradient control discriminates among models because it fixes a sign. A control against a smallest effect of interest adds a size criterion. For intention responses, the simulation's own intention-effect intervals (Section 4.2) supply the response test. Lin (2026b) recommends positive and negative controls, as does intended-use guidance (Abdurahman et al. 2025; Lin 2026a; Hullman et al. 2026). Equivalence and response controls bound different errors. Under the normal approximation, a TOST at the 5 percent level passes a simulation whose true simulated-minus-human difference is at least δ in absolute value with probability at most .05. A simulation with a true zero gradient passes the registered belief-gradient control (upper 95 percent limit below zero) with probability at most .025.

### 3. The Zero-Reference Check in 70 Survey Experiments

Ashokkumar et al. (2026) released GPT-4 predictions for 70 preregistered, nationally representative U.S. survey experiments. Applied to these predictions, the check shows which of GPT-4's equivalence passes occur where the human benchmark already lets the zero reference pass (exploratory, added after Study 1 unblinding). It uses the original authors' 156 hypothesized contrasts from 51 experiments and a ±0.20 margin in human reference-group SD, chosen for this archive from Cohen's (1988) small-effect benchmark. Of GPT-4's 58 equivalence passes at ±0.20 SD, 42 (72 percent [56, 86]) were also zero-reference passes. Across paired and independent prediction standard errors, total-SD units, and slope rescaling, the share was 68 to 72 percent (online supplement Table B2; human and simulated estimates are independent, Appendix B.2.2).

Under contrast weighting, the zero reference passed 46 percent of the hypothesized contrasts [34, 59] and unscaled GPT-4 predictions 37 percent [28, 46] (online supplement Table B3). Across all 1,678 pairwise contrasts, where the effect distribution, precision profile, and SD units change together, the pass rates fell to 18 and 13 percent. The share of GPT-4's passes that the zero reference also passed stayed at 72 and 70 percent. Experiments that supply more contrasts weigh more. Weighting each experiment equally brought the zero reference's pass rate to 39 percent on the hypothesized and 38 percent on all pairwise contrasts, so collection-level pass rates track which experiments supply the contrasts (online supplement Table B1).

The oracle reference reproduces each observed human estimate, and the nonresponsive noise reference has true effect zero. Both carry GPT-4's prediction noise. Conditional on the observed human estimates and GPT-4's prediction standard errors, 49.8 percent of the oracle reference's expected passes and 97.4 percent of the nonresponsive noise reference's are zero-reference passes. GPT-4's 72 percent was 22.6 points above the oracle reference's share (95 percent CI [13.1, 32.2]; online supplement Appendix B.2). The excess held at every margin from ±0.15 to ±0.30 SD (GPT-4's overlap 59 to 83 percent against the oracle reference's 37 to 67 percent; online supplement Table B4(a)). GPT-4's 58 passes fell short of the 61.4 expected from the nonresponsive noise reference (difference −3.4, 95 percent CI [−16.6, 9.0]; online supplement Table B4), and the 95 percent interval rules out an excess of more than 9 contrasts.

The benchmark sorts each contrast into one of three classes. In the first, the precision floor is at least δ and no independent prediction can pass. In the second, the zero reference passes. In the third, δ lies inside the informative range, where an equivalence pass can separate from the zero reference. At ±0.10 SD, Study 1's margin, the precision floor was at least 0.10 for 63 percent of the contrasts [47, 76]. At ±0.20 SD the classes held 0, 72, and 84. GPT-4 separated on 16 of the 84 contrasts in the third class, against 1.58 expected from the nonresponsive noise reference. The difference was 14.4 (95 percent study-cluster bootstrap CI [5.9, 24.0]; class counts at both margins in online supplement Table B4 and Appendix B.2.4). Across all sensitivity versions of the 156 contrasts, GPT-4's separations exceeded the expectation under the corresponding nonresponsive noise reference: 10 to 22 against 1.00 to 1.79 (online supplement Table B2). Choosing each contrast's margin after reading the predictions, rather than fixing δ at ±0.20 SD, raised the nonresponsive noise reference's expected number of contrasts with some margin of separation to 48.9, against 63 for GPT-4 (online supplement Table B4(e)). Figure 1 plots the 156 contrasts in the plane of the two thresholds.

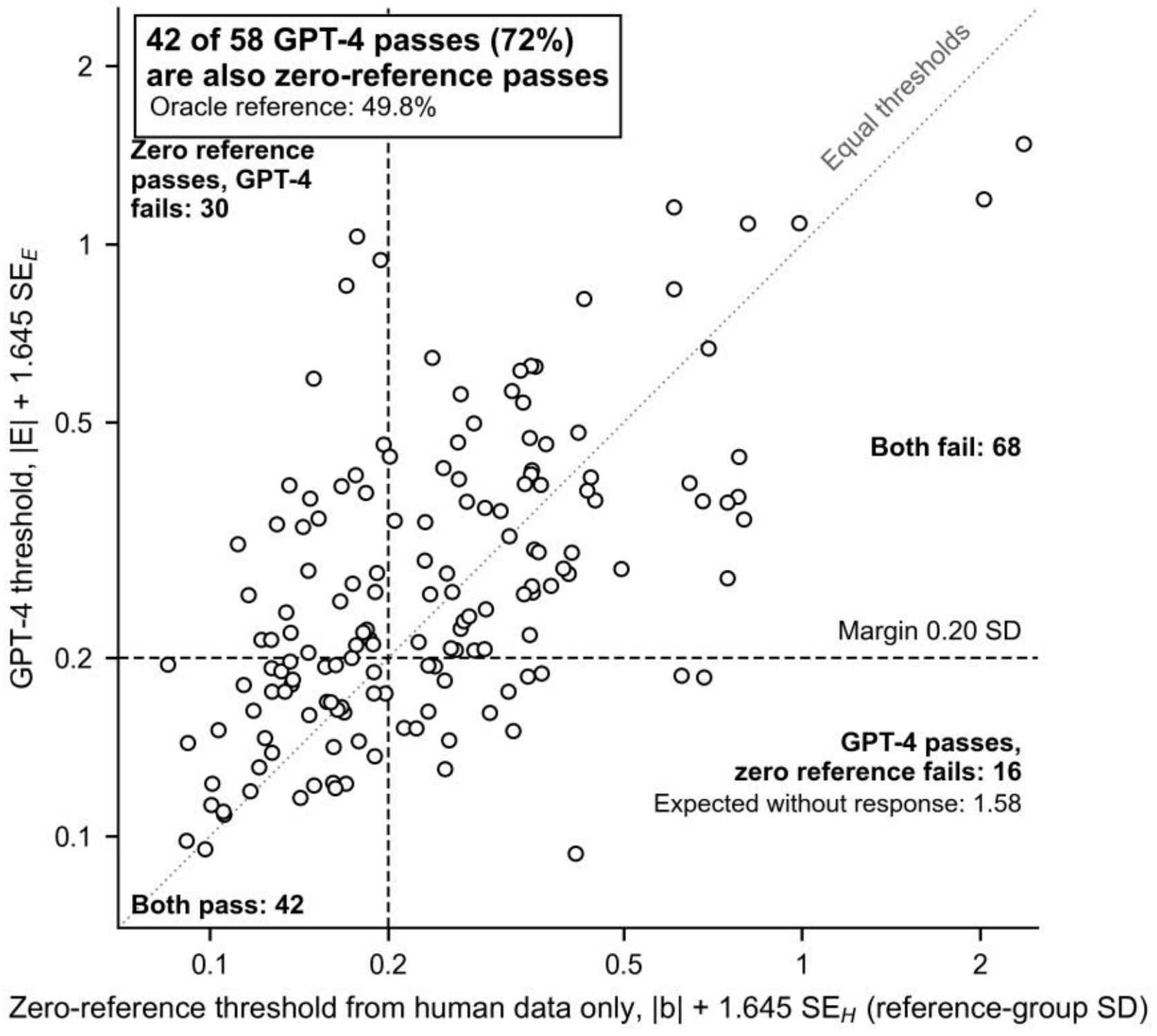

**Figure 1. The zero-reference threshold, computed from human data before simulation, sorts GPT-4's 58 passes at ±0.20 SD into 42 that are also zero-reference passes and 16 that separate from the zero reference.** Each point is one of the original authors' 156 hypothesized contrasts from 51 experiments. The x axis, labeled "Zero-reference threshold from human data only", is $|b_H| + 1.645\ SE_H$. The y axis, labeled "GPT-4 threshold", is $|E| + 1.645\ SE_E$ (unscaled predictions, paired prediction standard errors). Both are in human reference-group SD on log scales. The horizontal coordinate is available before simulation. The vertical coordinate, read against it, shows whether the completed simulation separates at the stated margin. The dashed lines mark the ±0.20 SD margin: GPT-4 passes below the horizontal line and the zero reference passes left of the vertical line. The two lines divide the plane into four groups, and the in-plot labels give their counts. The dotted diagonal marks equal thresholds: below it, some margin lets GPT-4 pass while the zero reference fails. The lower-right group, where GPT-4 passes and the zero reference fails, holds the 16 separations at ±0.20 SD, against 1.58 expected for a nonresponsive noise reference with GPT-4's paired prediction standard errors. An oracle reference reproduces every human estimate up to GPT-4's prediction noise. Conditional on the observed human estimates and GPT-4's prediction standard errors, about half (49.8%) of its expected passes at ±0.20 SD fall on contrasts that the zero reference passes, against 72% for GPT-4. Counts are contrast-weighted. Online supplement Table B2 gives the sensitivity versions. Status: exploratory, added after Study 1 unblinding.

The zero reference also sets the baseline for scores that need no margin. Leave-one-study-out slope rescaling raised GPT-4's MSE skill, the proportional reduction in mean squared error relative to the zero reference (Murphy 1988), from 0.10 to 0.66 and its passes at ±0.20 SD from 58 to 76. Of the 76, 54 (71 percent) remained zero-reference passes (online supplement Appendix B.2). Significance agreement is a two-sided 5 percent verdict that matches the human verdict. Its base rate is the zero reference's 93 of 156 contrasts (60 percent), the share of nonsignificant human results. Unscaled GPT-4 agreed with the human verdict on 91 contrasts with paired and on 110 with independent prediction standard errors.

In this archive, GPT-4's 58 passes fell 3.4 short of the 61.4 expected from the nonresponsive noise reference, the 95 percent interval rules out an excess of more than 9 contrasts, and its separations carried the response evidence.

## 4. Study 1: Effect Agreement and Response Controls in Two Social-Norm Experiments

In two registered experiments, equivalence, a response control, and response size give three separate verdicts on one simulation.

### 4.1. Design

Andre et al. (2026, wave 2) included 4,003 U.S. respondents. The treatment reported that 79 percent of Americans think people in the United States should try to fight global warming. The main outcome was a three-item index of public-expression intentions. Geiger et al. (2025, Study 1) included 3,361 climate-change believers in 11 countries. The treatment presented the country's consensus figure. The main outcome was intention to discuss climate change. Prior misperception is the pre-treatment norm estimate minus its benchmark truth, divided by the human control-group SD of the prior estimate. A gradient is the change in the outcome, in human outcome SD, per unit of prior misperception. Each respondent's pre-treatment pages and answers defined a persona, which keeps implied respondent characteristics fixed across arms (Gui and Toubia 2023). Each persona answered both arms in independent sessions, with two main-outcome replicates per arm averaged before analysis. Geiger used the original country-language materials, with an all-English version registered as a sensitivity analysis.

Claude Sonnet 5.5 (Sonnet), the registered primary model, returned a valid answer to all 50,906 scored calls (online supplement Table C3, panel F). Online supplement Appendix C.2 describes four open-weight panel models (Qwen, Gemma, Mistral, and GLM) and the OLMo exposure-audit model. Registered exposure probes addressed recall of a published result, a second route to agreement without response (Ludwig, Mullainathan, and Rambachan 2026). Sonnet recognized the Andre experiment in 20 of 20 calls. Truth recall, reproducing the published norm without a persona, was 0.00 on the formal channel, below the registered .50 threshold (online supplement Table C3, panels F and G). Online supplement Appendix C gives the questionnaire, material adaptations (Feuerriegel et al. 2026), and pilot exclusion rules. Online supplement Table A2 gives the access-channel deviation.

Study 1 compares source-specific coefficients in the registered regressions for the benchmark sample of respondents and their personas. Each regression stacks the observed human row and both simulated arm means with source interactions that let all coefficients differ. The target interaction estimates $d_{\mathrm{ATE}}$, the difference in average treatment effects (ATEs), or $d_{\mathrm{G}}$, the difference in gradients. These are the effect errors $E$, estimated on the human scale with CR1 standard errors that cluster all rows belonging to a person (Cameron and Miller 2015).

The confirmatory family contained four primary-model tests: $d_{\mathrm{ATE}} > 0$ and $d_{\mathrm{G}} < 0$ in each experiment, Holm-corrected at .05 (Holm 1979). Equivalence used TOST against ±0.10 human control-group SD (Baguley 2009), fixed before output. The gradient margin applies per unit of the prior misperception defined above. The registered belief-gradient control required the upper limit of the simulated gradient's 95 percent interval to be below zero. The registered "calibrated" label combined equivalence, a passing registered belief-gradient control, and a nondegenerate outcome under the registered endpoint-concentration and between-person-variation rules. The eight additional outcomes were preregistered as descriptive and stayed outside the Holm family. Online supplement Appendix A records decision timing, deviations, and analysis status (Willroth and Atherton 2024).

### 4.2. Effect Agreement and the Zero Reference

The registered question was whether the simulation overestimates the human effects. None of the four Holm-corrected directional tests (Section 4.1) rejected, and the two average-effect 95 percent intervals excluded overestimation above 0.08 SD. Both intention average effects passed equivalence and received the calibrated label, as did the Andre intention gradient (Table 1). The Geiger gradient did not receive the calibrated label. Its difference was −0.068, 95 percent CI [−0.127, −0.008]. Its one-sided $p$ was .013 before correction and .051 after Holm correction. Its 90 percent interval [−0.118, −0.018] extended beyond the equivalence margin. The registered English-material sensitivity analysis moved the Geiger average-effect difference but not its gradient difference (online supplement Table C3C).

**Table 1. Registered primary-model comparisons and decisions.**

| Experiment and intention-outcome quantity | Human coefficient (SE) | Simulated coefficient | Difference *d* [95% CI] | Holm *p* | Equivalence *p* | Registered calibrated label | Zero-reference threshold |
|---|---|---|---|---|---|---|---|
| Andre, average effect | 0.007 (0.028) | 0.028 | 0.021 [−0.034, 0.076] | .687 | .002 | Yes | 0.053 |
| Andre, gradient | −0.016 (0.028) | −0.002 | 0.014 [−0.041, 0.068] | .752 | <.001 | Yes | 0.061 |
| Geiger, average effect | 0.031 (0.030) | 0.040 | 0.009 [−0.049, 0.068] | .752 | .001 | Yes | 0.080 |
| Geiger, gradient | 0.036 (0.030) | −0.032 | −0.068 [−0.127, −0.008] | .051 | .144 | No | 0.085 |

*Note:* Registered alternatives: $d > 0$ for each average effect ($d_{ATE}$) and $d < 0$ for each gradient ($d_G$) in both experiments. None of the four rejected after Holm correction at .05. The registered across-experiment classification is "none" when neither experiment rejects, as for both average effects and gradients. Holm adjustment applies to the directional family. Equivalence decisions are per-cell tests under the registered rule (TOST at .05 against ±0.10, evaluated with 90% intervals); the table displays 95% intervals for the estimated differences. Units follow Section 4.1: average effects in human control-group outcome SDs, gradients in those SDs per unit of prior misperception. Differences use the registered stacked estimator with person-clustered standard errors (online supplement Appendix C.5 states the sampling frame). The calibrated label combines equivalence, a passing belief-gradient control, and a nondegenerate outcome (Section 4.1). No in the last row means the difference's 90% interval extended beyond ±0.10. The zero-reference threshold is $|b_H| + 1.645\ SE_H$ for the human coefficient and standard error in the same row. The zero reference passes the ±0.10 criterion when its zero-reference threshold is below 0.10, as in all four rows (column added after unblinding). Online supplement Table C3 gives the estimator and sensitivity results; Table C3I(a) gives the human standard errors and zero-reference quantities.

At the registered ±0.10, the human data alone showed a zero-reference pass on all four intention targets. Each of Sonnet's three equivalence passes matched a decision the benchmark had already fixed (Table 1; Figure 2a; online supplement Table C3I). The human estimates and standard errors were public when the margins were registered, so all four zero-reference thresholds (0.053 to 0.085) were computable at registration. Qwen and Gemma estimated negative Andre intention effects (−0.017 and −0.019; online supplement Tables C3A and C5) where the human estimate was 0.007. Both received the registered calibrated label. On that target the zero-reference threshold, 0.053, lies below the registered margin. The check therefore could have forecast at registration that an equivalence pass would match the zero reference's decision whatever the model's sign. Sonnet's own intention effects, 0.028 [0.023, 0.032] and 0.040 [0.032, 0.048] SD, lay above zero and below 0.05 SD (online supplement Table C3I(b), added after unblinding). These precisely estimated responses are smaller than any margin considered here. The human 95 percent intervals on these targets included zero (Table 1).

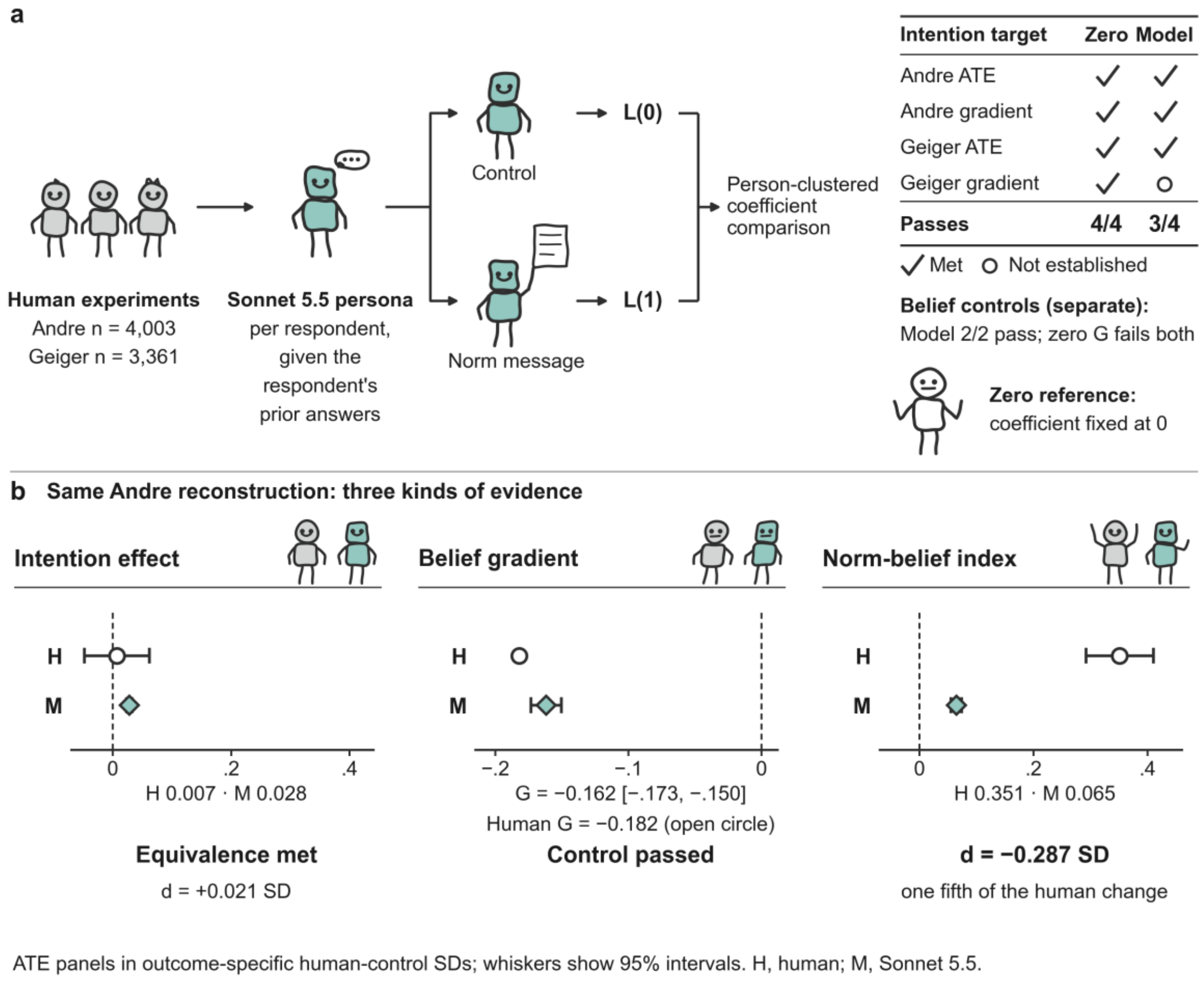

**Figure 2. Intention agreement, belief-gradient response, and mean belief change provide distinct evidence.** (a) Study 1 design and the intention targets. L(0) and L(1) are the persona's answers in the control and norm-message arms. The zero reference has a fixed zero coefficient with no estimation variance, so its decisions can be computed before simulation. The four intention equivalence targets and the separate belief-gradient controls test different quantities. The bottom row, "Passes", counts intention-equivalence passes out of four. The human data alone fix a zero-reference pass on all four targets (4/4), covering Sonnet's three passes (3/4). In the target table, the zero reference is evaluated on equivalence; below it, a zero gradient fails both belief-gradient controls. The margin of 0.10 is in the units of the corresponding ATE (average treatment effect) or gradient. (b) The same Andre reconstruction yields intention-effect equivalence and, on the same six-item action-specific norm-belief index, a passing belief-gradient control (−0.162 against −0.182 in humans). The index's mean change is 0.065 against 0.351 SD, one fifth of the human change. H and M denote human and Claude Sonnet 5.5 estimates. ATE whiskers show 95% intervals. Human n is 4,003 for intentions and 3,987 for beliefs, with outcome-specific human control SDs. The gradient panel shows the model's own 95% interval and the human point estimate as a hollow circle. G is measured in outcome SDs per unit of (prior minus truth)/human control SD(prior). Intention equivalence uses the registered, person-clustered difference and its 90% interval. Status: the norm-belief index effect is descriptive and was emphasized after inspection, and the zero-reference check was added after unblinding.

## 4.3. Response Controls and Norm-Belief Change

A persona that had underestimated support for climate action by more should raise its norm estimate by more after reading the page stating the true norm (Tankard and Paluck 2016; Sparkman, Geiger, and Weber 2022; Andre et al. 2024a). The belief gradient is therefore negative. Sonnet passed both registered belief-gradient controls (Figure 2a; the Andre gradient in Figure 2b). The simulated gradients were −0.162 [−0.173, −0.150] in Andre and −0.096 [−0.106, −0.087] in Geiger, against human gradients of −0.182 and −0.081. The control is a directional test that the zero reference fails, and it carries the response evidence fixed in advance by the design. GLM on Andre and OLMo on Geiger failed it. Qwen, Gemma, and Mistral passed on Andre with gradients of −0.032 to −0.045 against the human −0.182. The sign test and gradient size thus answer different questions (online supplement Table C3D).

On the six-item norm-belief index underlying the registered belief-gradient control, Sonnet's mean change was 0.065 SD against the human 0.351 (Figure 2b; descriptive, emphasized after inspection), about one fifth as large. Personas carry the respondents' own pre-treatment answers, so both sources have the same misperception distribution. A similar belief gradient (−0.162 and −0.182) can therefore coexist with a much smaller mean change. The norm-belief index has a zero-reference threshold of about 0.40 SD, so the zero reference fails at ±0.10. The difference's 90 percent interval, [−0.337, −0.237], lay entirely below −0.10. Sonnet's change was therefore smaller than the human change by more than the margin (difference −0.287). The margin was applied after unblinding (online supplement Tables C3J and C4). All five models that answered this outcome gave effects below 0.351 (from −0.075 to 0.197; online supplement Appendix C). In the human data the message moved norm beliefs while the intention effect stayed near zero (0.007). Norm-correction research documents this decoupling (Bursztyn and Yang 2022). Effect agreement, response direction, and response size address three different claims.

## 5. Study 2: Precision and Measurement in a Group Simulation

Before any group is simulated, the human benchmark fixes the smallest margin any independent group simulation can attain.

Study 2 reconstructed five-person hidden-profile discussions (the best option is visible only when members pool unshared facts; Stasser and Titus 1985; Lu, Yuan, and McLeod 2012) from the GRAIL (Group-AI Interaction Laboratory) experiment of Alsobay et al. (2026). The reconstruction used a 30-fact task (registered as exploratory; online supplement Part II). The human experiment randomized 70 groups per arm, including no facilitator (None) and an LLM facilitator (LLM). The facilitator effect on fact coverage was +2.9 facts, $SE_H = 0.81$, 95 percent interval [1.31, 4.49] (online supplement Appendix D.1). Fact coverage measures information surfaced in discussion, separately from decision quality (online supplement Appendix D.6). One model played all five members, and GPT-4o-2024-08-06 played the facilitator and produced the fact labels. The formal sample comprised 357 groups, and every attempted group was behaviorally valid and measurable (online supplement Appendix D). Qwen and Mistral ran 70 groups per arm, and Sonnet ran 38 None and 39 LLM groups (sample decision in online supplement Table A5). Gemma and GLM have no formal groups because the registered facilitator and annotator were unavailable on the continuation channel (online supplement Appendix A.2.4).

With 70 groups per arm, the precision floor, the smallest margin any independent simulation can pass, is $1.645 \times 0.81 = 1.33$ facts, almost half the 2.9-fact human effect. The zero-reference threshold is $2.9 + 1.645 \times 0.81 = 4.23$ facts, so the informative range of margins runs from 1.33 to 4.23 facts (Section 2.2).

Headroom, the outcome's upper limit minus the simulated control mean, caps any positive mean effect. Sonnet's None-arm mean was 27.24 of 30 facts, leaving 2.76 facts, less than the published human point effect of +2.9. Conditional on that control mean, the ceiling on $E$ against the point estimate is −0.14.

Against the human +2.9 facts, Sonnet's effect of +1.02 (own-effect interval [+0.51, +1.53]) gave error $E = -1.88$ (error interval [−3.55, −0.21]). Qwen's +1.39 (own-effect interval [+0.44, +2.33]) gave −1.51 (error interval [−3.36, +0.33]). Mistral's −1.79 under GPT-4o labels gave −4.69 (error interval [−6.48, −2.90]; Figure 3; planning calculations in online supplement Appendix D.10). All five registered sensitivity conditions kept Mistral's sign (−0.91 to −3.37; online supplement Appendix D.5). The error intervals' half-widths, 1.67, 1.85, and 1.79 facts, were close to the 1.59-fact minimum set by the benchmark's standard error.

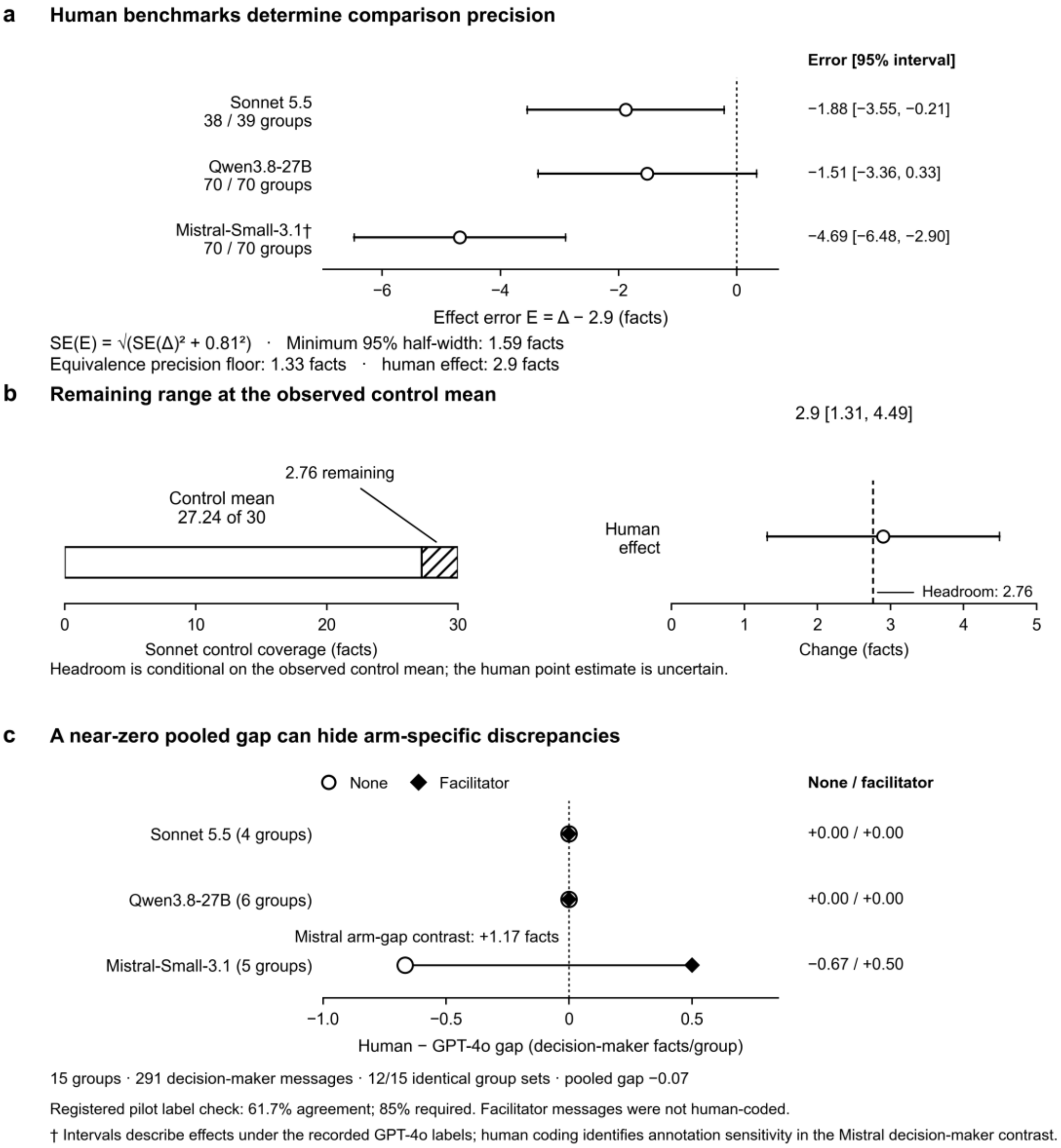
a
Human benchmarks determine comparison precision
Error [95% interval]
Sonnet 5.5
38 / 39 groups
−1.88 [−3.55, −0.21]
Qwen3.8-27B
70 / 70 groups
−1.51 [−3.36, 0.33]
Mistral-Small-3.1†
70 / 70 groups
−4.69 [−6.48, −2.90]
−6
−4
−2
0
Effect error E = Δ − 2.9 (facts)
SE(E) = √(SE(Δ)² + 0.81²) · Minimum 95% half-width: 1.59 facts
Equivalence precision floor: 1.33 facts · human effect: 2.9 facts
b
Remaining range at the observed control mean
2.9 [1.31, 4.49]
2.76 remaining
Control mean
27.24 of 30
0
10
20
30
Sonnet control coverage (facts)
Human
effect
Headroom: 2.76
0
1
2
3
4
5
Change (facts)
Headroom is conditional on the observed control mean; the human point estimate is uncertain.
c
A near-zero pooled gap can hide arm-specific discrepancies
None
Facilitator
None / facilitator
Sonnet 5.5 (4 groups)
+0.00 / +0.00
Qwen3.8-27B (6 groups)
+0.00 / +0.00
Mistral arm-gap contrast: +1.17 facts
Mistral-Small-3.1 (5 groups)
−0.67 / +0.50
−1.0
−0.5
0
0.5
Human − GPT-4o gap (decision-maker facts/group)
15 groups · 291 decision-maker messages · 12/15 identical group sets · pooled gap −0.07
Registered pilot label check: 61.7% agreement; 85% required. Facilitator messages were not human-coded.
† Intervals describe effects under the recorded GPT-4o labels; human coding identifies annotation sensitivity in the Mistral decision-maker contrast.

**Figure 3. Benchmark precision, control-arm headroom, and arm-specific measurement locate where a simulated group effect departs from the human effect.** (a) Effect errors $E = \Delta - 2.9$ facts, with 95% intervals combining independent simulation and human uncertainty, $SE_E = \sqrt{SE_\Delta^2 + 0.81^2}$. The human SE sets a minimum 95% half-width of 1.96 × 0.81 = 1.59 facts for error intervals, and a precision floor of 1.645 × 0.81 = 1.33 facts for equivalence margins. None/LLM group counts are 38/39 for Sonnet and 70/70 for Qwen and Mistral. (b) Conditional on Sonnet's observed None mean of 27.24/30, positive-change headroom is 2.76 facts. The separate change-scale display shows the human effect, 2.9 facts, with its 95% CI [1.31, 4.49]. (c) Human-minus-GPT-4o fact-count gaps by arm in the coded subsample. A paid research assistant, a PhD student blind to model and arm, evaluated 291 decision-maker messages from 15 groups: four Sonnet, six Qwen, and five Mistral, with two or three groups per model-arm cell. Connecting segments denote arm contrasts (descriptive; no intervals or tests). Within the five coded Mistral groups, the gaps are −0.67 in None and +0.50 in LLM (facilitator), a decision-maker-only difference of 1.17 facts beside a pooled gap of −0.20 facts. Across all 15 coded groups, the pooled gap is −0.07 facts, and 12/15 group fact sets agree exactly. Intervals describe effects under the recorded GPT-4o labels; the 1.17-fact arm difference exceeds the registered half-fact reference. Online supplement Appendix D gives the full measurement results, including the Mistral arithmetic (D.8). Status: Study 2 is exploratory, and human coding was planned after unblinding and before coding.

Under the separation condition of Section 2.2 (post-unblinding arithmetic; online supplement Table A7), the margins at which a model separates from the zero reference are (3.28, 4.23] facts for Sonnet and (3.07, 4.23] for Qwen. Both windows lie above the 2.9-fact human effect, so in this benchmark separation up to $\delta_{.05}$ confirms the direction of the facilitator effect and leaves its size to the error interval. At $\delta = 3.5$, an illustrative margin inside both windows, $P_0$ is 0.001 for Sonnet and 0.024 for Qwen. Because $\delta_{.05}$ (3.87 and 3.66) exceeds 3.5, separation there is itself a 5 percent test of response, and both own-effect intervals exclude zero (online supplement Table D3(c)). Mistral's simulation threshold, 6.19, exceeds the zero-reference threshold, 4.23, so no margin separates it.

Coverage counts distinct facts per transcript. The registered pilot check required 85 percent per-message exact agreement between GPT-4o and an isolated LLM labeler. Agreement reached 61.7 percent, and under the registered branch for a failed check the first author retained the GPT-4o measure (online supplement Table A4). On all five pilot transcripts, GPT-4o's facts were a subset of the labeler's, and across four reference annotations each formal effect changed by at most 0.23 facts, within the registered half-fact bound (online supplement Appendix D.7).

Arm-specific annotation gaps determine error in the arm contrast. Annotation errors that vary with the coded arm can bias downstream estimates (Ashwin, Chhabra, and Rao 2026), so a measurement check assesses error at the unit that builds the arm contrast (Flake and Fried 2020). In a human audit, added after Study 2 unblinding and specified before coding, a paid research assistant (a PhD student blind to model and arm) coded 291 decision-maker messages from 15 formal groups. The pooled human-minus-labeler gap was −0.07 facts, and 12 of 15 group fact sets matched: all ten Sonnet and Qwen groups and two of the five Mistral groups. For decision-maker-message coverage in the five coded Mistral groups, three one-fact discrepancies gave a 1.17-fact arm difference, above the registered half-fact reference, beside a pooled gap of −0.20 facts (descriptive). Online supplement Appendix D.8 gives the arithmetic and the facilitator-message scope.

Precision, headroom, and arm-specific measurement each limit what a group audit can establish.

## 6. A Protocol for Benchmark Checks

Report the zero-reference threshold beside every equivalence decision. The check changes the order of validation: assess what the benchmark can distinguish before allocating simulation effort, then add the controls the intended claim requires. A local average effect can rely on the equivalence decision under its stated uncertainty. A treatment-response claim needs the model's own effect and a directional control on the outcome and statistic that the claim names, as in Study 1. Planning needs the effect and its variance sources stated together. Automated measurement needs a check on the arm contrast. In Study 2, a use that requires a margin below 1.33 facts calls for a more precise human benchmark (Section 5). Table 2 gives the five steps and their readings. Online supplement Appendix E.2 gives the Python and R implementations for independent and joint comparisons, with worked examples.

**Table 2. Benchmark checks as a five-step protocol, with their readings in the archive and the two studies.**

| Step | Computation and decision | Reading |
|---|---|---|
| 1. Specify the use, target statistic, unit, and margin δ | Name the quantity the claim concerns and the model configuration evaluated; justify δ in the quantity's units in advance; the use's decision loss bounds δ above (Lakens, Scheel, and Isager 2018), and prespecifying δ gives separation its size (online supplement Appendix B.2.4) | Study 1: four intention targets at ±0.10 human control-group SD (Section 4.1) |
| 2a. Before simulation, identify whether the estimates are independent or jointly estimated, and compute the zero-reference threshold $\lvert b_H \rvert + 1.645\ SE_H$ and, for independent estimates, the precision floor $1.645\ SE_H$ | Independent comparisons continue to 2b; joint comparisons continue to step 3 (scope in the note) | Study 2: zero-reference threshold 2.9 + 1.645 × 0.81 = 4.23 facts and floor 1.645 × 0.81 = 1.33 facts (Section 5) |
| 2b(a). Independent comparison: δ at or below the floor | No independent prediction can pass | Archive: 0 of 156 contrasts at ±0.20 SD (Section 3) |
| 2b(b). Independent comparison: δ above the floor and at or below the zero-reference threshold | A sufficiently precise simulation can pass while the zero reference fails, so read the simulation threshold; with a planned $SE_E$, compute $P_0$ and $\delta_{.05}$ before simulation | Archive: 84 contrasts at ±0.20 SD; Study 2: margins from 1.33 to 4.23 facts (Sections 3 and 5) |
| 2b(c). Independent comparison: δ above the zero-reference threshold | The zero reference passes, so equivalence records agreement within δ and claim-specific controls (step 4) assess response | Archive: 72 contrasts at ±0.20 SD (Section 3) |

| Step | Computation and decision | Reading |
|---|---|---|
| 3. After simulation, compute $\|E\| + 1.645\ SE_E$ with the dependence structure of the comparison, and report separation at δ | The margin fixed in step 1 governs the decision. For an independent comparison, report $P_0$ and $\delta_{.05}$ at the realized $SE_E$, updating any planned values; up to $\delta_{.05}$, separation is a 5 percent test of response (Section 2.3), and when δ exceeds $\|b_H\|$, the zero reference fails through $SE_H$ alone, so read the simulation's own interval for response | Archive: at ±0.20 SD, GPT-4 separated on 16 of the 84 informative contrasts against 1.58 expected from the nonresponsive noise reference (Section 3; Figure 1); Study 2 margins of separation in Section 5 |
| 4. Add claim-specific controls: response, headroom, and arm-contrast measurement | Controls address the outcome and statistic the claim names: directional response controls need simulated outcomes, headroom a measured control arm, and arm-contrast measurement suitable reference coding | Study 1: belief-gradient control passed (−0.162), mean change 0.065 SD against 0.351 SD (Section 4.3); Study 2 headroom and arm-contrast measurement in Section 5 |
| 5. Report all outcomes and analysis statuses | State the scale and margin, then display every registered outcome, interval, and analysis identity | Study 1: all ten registered outcomes, each with its analysis status (online supplement Table C4) |

*Note:* The informative range of margins (Section 2.2) runs from the independent-comparison precision floor, the smallest margin an independent prediction can pass, to the zero-reference threshold, the largest margin at which the zero reference fails. Its width is $|b_H|$. The floor, the informative range, $P_0$, and $\delta_{.05}$ use the independent-comparison variance, and a planned $SE_E$ gives $P_0$ and $\delta_{.05}$ before simulation. A joint comparison, such as a stacked estimator on shared respondents (Study 1), uses the zero-reference and simulation thresholds and the separation decision. Analysis statuses appear in the Preregistration Statement and online supplement Table A7.

## 7. Discussion: Implications for Validation Practice

### 7.1. What Simulation Validation Can Read from the Benchmark

Before simulation, the human benchmark determines whether equivalence at the chosen margin can distinguish a simulation from the zero reference. An equivalence pass supports agreement within that margin. Response direction, response size, and transport to another use each require evidence for the corresponding claim. In an independent comparison, the precision floor, 1.645 $SE_H$, belongs to the human study. A use that needs a margin below it therefore needs a more precise benchmark (Sections 3 and 5). Failure to separate can reflect a benchmark with little power to verify response, even when the simulated effect is correct (Section 2.3).

Pass counts record benchmark composition. Response evidence lies in separations at a margin fixed in advance (Section 3). Pass rates and scores are read beside the zero reference's. Aggregate replication rates are read with each collection's margin, precision profile, and zero-reference pass rate (Cui, Li, and Zhou 2025; J. S. Park et al. 2026). The replication literature separates replication rates from false-positive prevalence (Ben-Menachem, Galper, and Fishman 2026). Forecasts of experimental results (DellaVigna and Pope 2018; Ashokkumar et al. 2026) are scored the same way, as predictive models are read beside simple baselines (Murphy 1988; Hofman, Sharma, and Watts 2017; Zheng and Cheng 2025).

Discrimination across contrasts is a further claim with its own evidence, as are properties beyond the average, such as issue-specific bias in survey simulations (Boelaert et al. 2025). A positive rescaling leaves the correlation between predicted and human effects unchanged, yet slope rescaling raised GPT-4's MSE skill (Section 3; online supplement Appendix B.2.3). Compressed variance (Bisbee et al. 2024; P. S. Park, Schoenegger, and Zhu 2024), homogeneous generated text (Zhang, Xu, and Alvero 2025), and exaggerated regularities (Aher, Arriaga, and Kalai 2023) motivate separate checks on response size (Gelman and Carlin 2014) and dispersion. Standardizing by the simulated answers' own SD defines a different target (online supplement Table C3, panel E).

### 7.2. Relation to Replication Assessment, Negative Controls, and Prediction-Assisted Inference

The claims above cover stand-ins for social agents (Dillion et al. 2023), simulated samples (Argyle et al. 2023), and forecasts of experiments (Lippert et al. 2024). They also cover economic decisions (Filippas, Horton, and Manning 2024; Tranchero et al. 2026) and group deliberation (Tessler et al. 2024; Qian et al. 2025; Tsirmpas et al. 2026; Tak et al. 2026; Jang et al. 2026; Parisi et al. 2026; Toxtli and Delaflor 2026; Li, Naito, and Shirado 2026). Agent-based modeling has a longer tradition of confronting simulated social processes with data (Bruch and Atwell 2015; Banisch and Shamon 2025).

The zero reference is an analytic baseline for what the equivalence test can discriminate, whereas an empirical negative control tests assumptions about the structure of bias (Lipsitch, Tchetgen Tchetgen, and Cohen 2010). The thresholds apply the established idea of inverting an equivalence test into a minimum supported margin (Hartman and Hidalgo 2018). Applied to a fixed zero prediction, the inversion gives a benchmark check available before simulation, and the separation calculation characterizes its response-test interpretation. Replication assessment compares two noisy estimates of one effect with criteria that depend on the original estimate's precision (Simonsohn 2015; Anderson and Maxwell 2016; Patil, Peng, and Leek 2016; Hedges and Schauer 2019). Assay sensitivity in noninferiority trials rests on a constancy assumption about the control's effect (Temple and Ellenberg 2000; D'Agostino, Massaro, and Sullivan 2003), whereas the zero-reference threshold uses the observed benchmark.

PPI (Angelopoulos et al. 2023) and its mixed-subjects form (Broska, Howes, and van Loon 2025) combine predictions with human observations through unit-level predictive information. They provide PPI correlation, effective sample size, and power analysis for planning. Design-based supervised learning (DSL; Egami et al. 2023; Rister Portinari Maranca et al. 2025) uses reference annotations collected under a stated sampling design to correct automated measurement. The zero-reference check addresses a third task: using a benchmark's published estimate, standard error, and margin to show what an equivalence decision can distinguish.

### 7.3. A Research Agenda for Benchmark Design

Benchmark design must jointly address discrimination at decision-relevant margins and representation of the settings in which simulations will be used. Anthis et al.'s (2025) iterative evaluation program for LLM social simulation can incorporate four questions of benchmark design.

First, benchmark composition decides what a collection can validate (Section 3). Near-zero benchmarks remain useful for validating negligible effects. Collections intended to validate response also need contrasts on which the zero reference fails. Collections can report the zero-reference threshold and precision floor for every prespecified contrast, and pooling comparable replications (Hedges and Schauer 2019) can lower the floor. How should benchmark collections balance discrimination with representation?

Second, response and measurement claims each need their own controls, such as the registered belief-gradient control (Section 4.3) and the arm-specific coding audit (Section 5). Codebook and coder comparisons assess the scoring procedure (Than et al. 2025; Sturgis et al. 2026; Halterman and Keith 2026; Stuhler, Dang Ton, and Ollion 2025; Gilardi, Alizadeh, and Kubli 2023). Arm-specific measurement error biases causal contrasts built on learned proxies (Knox, Lucas, and Cho 2022). Connecting fact coverage to deliberative information sharing is a measurement-validity argument of its own (Adcock and Collier 2001). Sampling and annotation designs that make arm-contrast error estimable at the unit used to construct the treatment effect are part of this agenda. Matched sham messages with a human benchmark can extend response controls, and a joint-estimator $P_0$, as in Study 1, would carry the response test to stacked designs. Which controls distinguish sensitivity to intervention content from sensitivity to wording or numerical format? Third, margins need a stated origin. The margin's position in the informative range (Section 2.2) shows when a use needs a more precise benchmark, a different target quantity, or no equivalence claim. Applications specify a substantively defensible tolerance before simulation (Hartman and Hidalgo 2018; Lakens, Scheel, and Isager 2018). Deriving that tolerance from decision loss is a research task (online supplement Appendix C.5; Manski 2019). How should margins reflect decision consequences when errors have different costs across interventions or populations? Fourth, validation transport requires explicit links among the benchmark, population, model configuration, and intended use (Table 2, step 1; Egami and Hartman 2023). Model and configuration choices move LLM results (Chung 2026; Cummins 2026; Lyman et al. 2025), as English materials did for the Geiger average effect (Section 4.2). The empirical scope here (Simons, Shoda, and Lindsay 2017) is effect-level evidence from U.S. survey experiments, two social-norm experiments, and a group task. Representation of a population's opinions (Santurkar et al. 2023) and portrayal of identity groups (Wang, Morgenstern, and Dickerson 2025) are separate validation targets. Transport to a proposed use also needs temporal assumptions and a repeated check after a model update (Munger 2023). Which changes in population, materials, model version, or configuration require a new benchmark comparison?

### 7.4. Conclusion

Before any model runs, a benchmark's estimate, standard error, and margin show whether the zero reference passes the equivalence criterion. In a 70-experiment archive the zero reference passed 42 of the 58 contrasts that GPT-4 passed at ±0.20 SD (72 percent). Where it failed, GPT-4's 16 separations exceeded the 1.58 expected without response. Within one experiment, matched intention effects, a passing belief-gradient control, and a norm-belief change about one fifth of the human change addressed three different claims. The group audit adds a 1.33-fact precision floor against a 2.9-fact effect, headroom, and arm-contrast measurement. Before simulation, the benchmark check tells each validation which claim its agreement can support and what evidence each further claim needs. Report the zero-reference threshold beside every equivalence decision.

### Preregistration Statement

Study 1 was registered on OSF Registries before pilot output and again before formal calls, and its four directional comparisons and calibration rules stayed fixed. Study 2 was registered before group simulation, and its updates recorded the continuation decision after the pilot annotation check (online supplement Table A4), the final sample, and the human-coding plan. The plan followed Study 2 unblinding and preceded coding. The single-agent follow-up was registered before its own trials, after Study 2 results were known. Online supplement Appendix A reports all deviations, decision timing, and results available at each decision, with each analysis identified in Table A7. Part II reproduces the English registration texts and updates. The complete Chinese Study 2 specification remains in the archived registration. The links to the time-stamped registrations are masked for anonymized review and will be added at conditional acceptance.

Analyses added after unblinding are labeled where first reported (Lakens 2024; Nosek et al. 2018) and are of two kinds. The empirical ones are the zero-reference check with separation at $\delta$, the descriptive equivalence classifications, and the archive application, all added after Study 1 unblinding. Planning calculations, process analyses, and stance trajectories retain their stated post-unblinding status. The archive application, including its significance-agreement rates, oracle-reference shares, the nonresponsive noise reference with its pass-rate comparison, and skill scores, is exploratory, with margins chosen retrospectively for that archive. The analytic ones are the scaling corollary, the dominance proposition, the nonresponse separation probability, the informative-range rule with its sample-size form, $\delta_{.05}$, and the joint operating characteristics of online supplement Table B7. Both $\delta_{.05}$ and Table B7 were added after unblinding, as was the step-3 arithmetic for Study 2, which registered no margin.

**Code Materials and Data Availability**

The online supplement provides protocols, complete reported results, and a computational file and field index. The replication package contains the code and materials used to generate and analyze the simulations, together with frozen aggregate outputs. Run offline, the package rebuilds eight table-level summaries and the arithmetic diagnostics from those outputs. Appendix E maps each reproduction route, including record-level reestimation and regeneration of model outputs. Study 1 uses the original authors' publicly available, de-identified data. Study 2 uses published GRAIL statistics and materials; no GRAIL participant-level records were used. The archive application uses the original authors' public Code Ocean capsule (Hewitt et al. 2026; data CC0; https://codeocean.com/capsule/9843791/tree/v1) and calls no model. The package refits that analysis from the capsule, including its significance-agreement rates. It also contains benchmark_check.py, which computes steps 2 and 3 of the Section 6 protocol, and operating_characteristics.py, which produces Figure B1 and Table B7 (online supplement Appendix E.2). The archived registration materials identify the frozen simulation and analysis specifications. On the registered full-precision results file, two independently written implementations agree to within a maximum relative difference of $1.7 \times 10^{-12}$ (online supplement Appendix C.5). The study records retain prompts, model identities, access channels, sample decisions, and analysis timing. Before acceptance, the data, code, and materials, including the shareable simulated analysis records and the complete analysis code, will be made openly available in Figshare under a persistent DOI. Study 1 uses the Andre et al. (2026) replication data (Andre et al. 2024b; Harvard Dataverse, doi:10.7910/DVN/CEFHPP) and the Geiger et al. (2025) cleaned data (Geiger 2025; OSF project r2byz). The archive application uses that Code Ocean capsule (Hewitt et al. 2026).

**Institutional Review and Consent**

This study was reviewed and approved by the first author's institution. The behavioral comparisons use publicly available, de-identified data and published aggregate statistics. No new human behavioral experiment was conducted for this article. A paid research assistant coded simulated decision-maker messages, which contain no human participant responses.

**Acknowledgments**

This work was supported by the AMD University Program (AUP). The computations were partly performed using the Genkai supercomputer under the program of the Research Institute for Information Technology, Kyushu University, Japan. We thank the research assistant who coded the simulated transcripts.

**Funding**

This work was supported by MEXT Supporting Pioneering Research through AI for 1,000 Discovery challenges Program (SPReAD) Japan Grant Number JPMXP1726306059.

**Declaration of Conflicting Interests**

The authors declared no potential conflicts of interest with respect to the research, authorship, or publication of this article.